\documentclass[conference]{IEEEtran}
\usepackage{hyperref}
\usepackage{graphicx} 
\usepackage{caption} 
\usepackage{cite} 
\usepackage{graphicx}
\usepackage{algorithm}
\usepackage{algpseudocode}
\usepackage{xcolor}
\usepackage{balance}
\usepackage{amsmath} 
\usepackage{amsmath,amssymb}
\usepackage{multirow}
\usepackage[T1]{fontenc}
\IEEEoverridecommandlockouts                              

\usepackage{mathptmx} 
\usepackage{amsmath} 
\usepackage{amssymb}  
\usepackage{xcolor}
\usepackage{soul}
\usepackage{rotating}

\begin{document}









\title{Self-Closing Suction Grippers for Industrial Grasping via Form-Flexible Design}

\author{Huijiang Wang$^1$,~\IEEEmembership{Student Member,~IEEE,}
        Holger Kunz$^2$,\\
        Timon Adler$^2$,
        and Fumiya Iida$^{1}$,~\IEEEmembership{Senior Member,~IEEE}%

\thanks{$^{1}$Bio-Inspired Robotics Lab, Department of Engineering, University of Cambridge, Cambridge CB2 1PZ, United Kingdom.}
\thanks{$^{2}$FORMHAND Automation GmbH, 38118 Braunschweig, Germany.}

\thanks{\tt\small Emails: hw567@cantab.ac.uk, h.kunz@formhand.de}}

\maketitle

\begin{abstract} 

Shape-morphing robots have shown benefits in industrial grasping. We propose form-flexible grippers for adaptive grasping. The design is based on the hybrid jamming and suction mechanism, which deforms to handle objects that vary significantly in size from the aperture, including both larger and smaller parts. Compared with traditional grippers, the gripper achieves self-closing to form an airtight seal. Under a vacuum, a wide range of grasping is realized through the passive morphing mechanism at the interface that harmonizes pressure and flow rate. This hybrid gripper showcases the capability to securely grasp an egg, as small as 54.5\(\%\) of its aperture, while achieving a maximum load-to-mass ratio of 94.3.

\end{abstract}

\section{Introduction}

Inspired by natural organisms, many shape-changing robots have been proposed. Utilizing soft materials, soft robotic grippers perform tasks where traditional rigid grippers suffer such as adaptive manipulation and safe handling of delicate objects \cite{wang2017prestressed}. Soft grasping can be categorized into three approaches \cite{shintake2018soft}: actuation, controlled stiffness and controlled adhesion. Actuation and controlled adhesion methods typically involve control systems for motors or soft materials \cite{wang2023self}. Recently, research in soft grasping has shifted toward performing grasping tasks while minimizing control complexity \cite{gilday2020suction}. 



\textcolor{black}{Grasping mechanisms have evolved from traditional finger-based pinching \cite{deimel2016novel} to more underactuated methods including universal grasping approaches \cite{amend2012positive,hwang2023self} and suction-based techniques \cite{kim2023octopus}. Table \ref{tab:graspModes} lists the comparison of these grasping mechanisms. The pinching can be achieved by cable-driven, pneumatic fingers and layer jamming. These grippers rely on friction upon contact and the pinching action between two or more soft fingers. Universal grippers form friction or interlocking with the object \cite{brown2010universal} based on a morphological transition from a soft unjammed state to a stiffened jammed state that occurs by adjusting the gripper's stiffness. However, both face challenges with bulky objects because they rely on wrapping up the object. In contrast, by attaching to a small area on the object, suction cups require only simple control. They generate high handling (suction) force but the adhesion quality is sensitive to object texture and geometric irregularity.}


\begin{table}[ht]
\centering
\caption{\textcolor{black}{Handling modes and the comparison.}}
\label{tab:graspModes}
\begingroup
\fontsize{10pt}{14pt}\selectfont
\resizebox{\columnwidth}{!}{%
\begin{tabular}{|p{1.2cm}|p{3.5cm}|p{3.5cm}|p{1.5cm}|p{1.5cm}|}
\hline
\textbf{Mode} &
  \textbf{Handling modality} &
  \textbf{Mechanism} &
  \textbf{Object size\textsuperscript{*}} &
  \textbf{Lift ratio\textsuperscript{**}} \\ \hline
\multirow{3}{*}{\rotatebox{90}{\parbox{1.5cm}{\centering Single Mode}}} &
  Fingered Gripper (F) \cite{deimel2016novel,wang2017prestressed} &
  Pinching (P) &
  $\leq$ &
  + \\
 &
  Universal Jamming Gripper (UJG) \cite{amend2012positive,brown2010universal} &
  Friction (FR), Interlocking (LK) &
  $<$ &
  ++ \\
 &
  Suction Cup (SC) \cite{kim2023octopus} &
  Suction (S) &
  $\geq$ &
  +++ \\ \hline
\multirow{5}{*}{\rotatebox{90}{\parbox{1.5cm}{\centering Hybrid Mode}}} &
  UJG + SC \cite{gilday2023xeno,fang2022soft,krahn2017soft} &
  S + FR &
  $<$ &
  + \\
 &
  F + SC \cite{chin2020multiplexed,li2023bioinspired} &
  P + S &
  $<$ &
  + \\
 &
  F + UJG \cite{lee2021soft,fang2022multimode} &
  P + FR &
  $<$ &
  + \\
 &
  F + UJG + SC \cite{washio2022design} &
  P + FR + S &
  $\leq$ &
  ++ \\ 
 &
  \textbf{This work} &
  \textbf{S + FR} &
  \textbf{$\leq$, $\geq$} &
  \textbf{+++} \\ \hline
\end{tabular}%
}%
\endgroup
\begin{flushleft}
\textsuperscript{*} Object size: Applicable object size compared to the gripper's aperture. \\
\textsuperscript{**} Lift ratio: Load-to-mass ratio with three-scale qualitative evaluation.
\end{flushleft}
\end{table}

Prototypes that combine fingers and suction cups use a movable sucker to assist gripping when the fingered mode alone cannot complete the task \cite{hao2020multimodal}. A gripper combining finger pinching and universal gripping achieves multimodal grasping with varying palm stiffness \cite{lee2021soft} but is challenged by fault tolerance. By putting the fingers into the jammed soft cover, Washio et al. \cite{washio2022design} proposed a gripper design with two pneumatic fingers into a jamming chamber to actively adjust the aperture size. However, this hybrid gripper has limited holding force due to the constrained deformability of the pneumatic fingers. Gilday et al. \cite{gilday2023xeno} introduced a gripper with enhanced adhesion, but its operable object size is limited by the aperture dimensions due to the gripper's "\(\cap\)"-shaped cross-sectional design.

Suction cups require secure attachment to combat air leaks. Existing designs suffer from porous materials and complex textures. This paper introduces a form-flexible design capable of grasping various objects in real-world industrial environments. \textcolor{black}{The hybrid design is validated to enhance grasping performance by addressing the limitations of single jamming or suction modes.} The sealing lip enables an airtight closure at the interface. The gripper is suitable for heavy-load scenarios including texture grasping, agriculture, and construction. The form-flexible design has been physically validated in industrial-focused grasping ranging from delicate objects like eggs to heavy, high-roughness, and bulky objects such as construction bricks and car assembly parts. \textcolor{black}{As such this paper makes a number of contributions:}

\begin{itemize}
    \item     \textcolor{black}{Shape-morphing grippers with hybrid jamming and suction mechanisms are designed to grasp objects smaller than its aperture and to handle heavy loads.}
    
    \item     \textcolor{black}{Development of a sealing lip that achieves self-closure for enhanced adhesion during grasping.}

    \item     \textcolor{black}{Proof of reliability of three representative form-flexible grippers in various real-world industrial environments.}
    
\end{itemize}


\section{Form-Flexible Grippers}

\subsection{Gripper Design}

\begin{figure}[h]
    \centering
    \includegraphics[width=1\columnwidth]{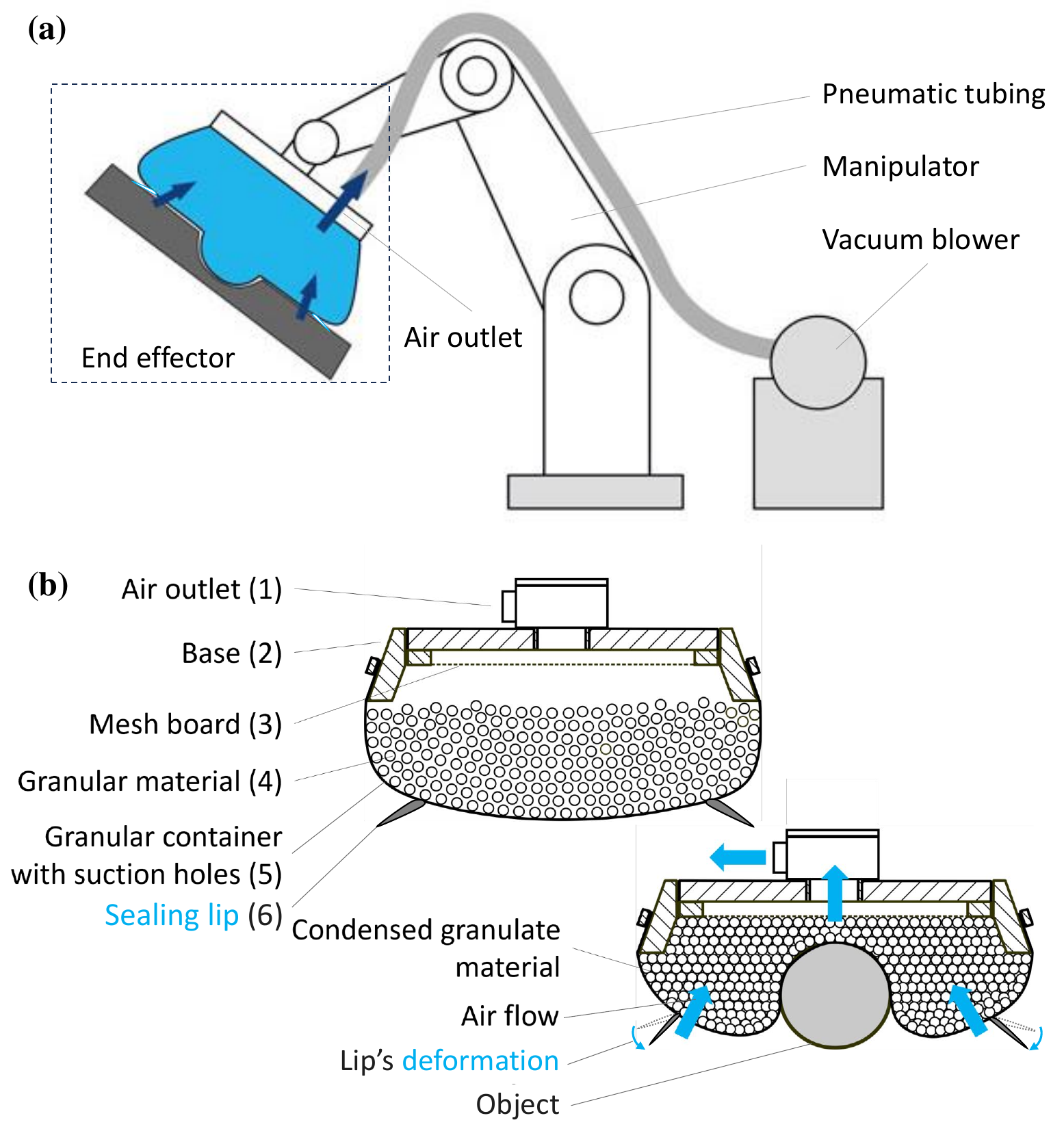}
    \caption{The working principle of the form-flexible grasping. (a) The gripper is powered by a vacuum blower for stiffness transition. A robotic arm is used for spatial orientations. (b) The chamber is not fully filled with granules until they flow and squeeze to the base. There is a constant airflow (blue arrows) through condensed granules during suction. \textcolor{black}{This airflow induces deformation in the sealing lip, which enhances the adhesion and grasping.} }
    \label{fig:structure}
\end{figure}

The robotic system comprises a gripper as the end-effector, a robotic manipulator, a vacuum blower as the power source \textcolor{black}{and the pneumatic connection with an air supply system.} The gripper features a flexible outer layer that adapts to various object shapes \textcolor{black}{with its internal content capable} of transitioning between fluid-like and rigid states. The hybrid jamming and suction design (Figure \ref{fig:structure}) includes a rigid base, a mesh board to prevent granule suction and a polyurethane soft \textcolor{black}{interface for deformability.} Encapsulated polymer granules are housed in a porous container and \textcolor{black}{all components} are secured with bolts.

FormHand has commercialized this form-flexible grasping technology \textcolor{black}{and offers products including} the FH-R80, FH-R150 and FH-E3020 with customizable dimensions \cite{lochte2014form}. These grippers are easily interchangeable \textcolor{black}{on the robotic manipulator to meet different industrial needs} and require minimal control complexity. Operable by a human user or robotic arm, the \textcolor{black}{end-effector} is controlled via a simple vacuum pressure valve switch.



\subsection{Shape-Morphing Model}

\begin{figure}[b]
    \centering
    \includegraphics[width= \columnwidth]{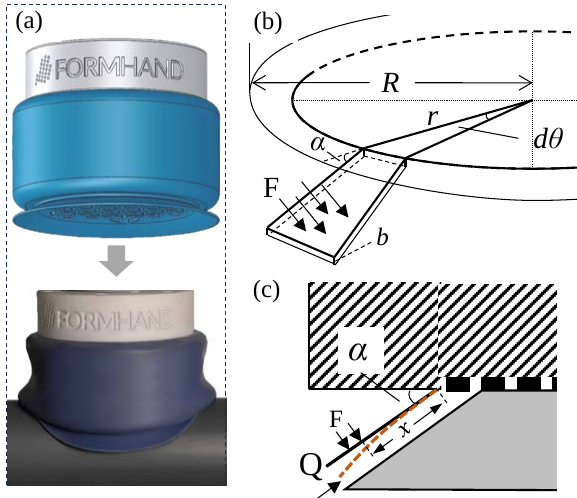}
    \caption{The deformation of the sealing lip under the airflow.}
    \label{fig:morphModel}
\end{figure}

We use the infinitesimal dividing modeling method to represent the deformation of the sealing lip. As depicted in Figure \ref{fig:morphModel}b, the small segment can be considered as an isosceles trapezoid with the median as \(\frac{1}{2}\left( r+R \right)d\theta\) and the height as \(\frac{R-r}{cos\alpha}\). Therefore the area of this segment is:

\begin{equation}    \label{eq:area}
    \textcolor{black}{dA=\frac{1}{2}\frac{R^2-r^2}{cos\alpha}d\theta}
\end{equation}

When the vacuum is activated, the \textcolor{black}{high-speed airflow within the channel produces a region of reduced pressure inside the channel, which is enclosed by the sealing lip} and the object (\(P_{in}\)). This pressure \textcolor{black}{differential relative to the external environment (\(P_{air}\))} can be explained through Bernoulli's principle.

\begin{equation}
    P_{air} - P_{in} = \textcolor{black}{\frac{1}{2}\rho\left( \left( \frac{Q}{A} \right)^2 - v^2_{out} \right)}
\end{equation}

where \(Q\) \((m^3/h)\) denotes the airflow rate and \(\rho\) is the air density \((kg/m^3)\). \textcolor{black}{The external airflow velocity \(v_{out}\) is assumed to be zero. The resulting pressure difference generates a self-closing force on the sealing lip, which is the product of the pressure difference and the area \(F=\Delta P \times A\)}. The resulting pushing force on a segment is expressed as:

\begin{equation}   \label{eq:areaForce}
    F = \textcolor{black}{\frac{\rho Q^2 cos\alpha}{\left( R^2 - r^2 \right)d\theta}}
\end{equation}

We assume that the force applied to this segment is uniformly distributed onto the area. The segment is assumed to be a cantilever beam. Consider the Modulus of elasticity of the material of the sealing lip as \(E\) \((Pa) \). As depicted in Figure \ref{fig:morphModel}c, we have the beam deflection under a uniformly distributed load.

\begin{equation}   \label{eq:materilMechanics}
    y = \frac{Fx^2}{24EI}\left(x^2 + 6L^2-4Lx \right)
\end{equation}

where \(y\) is \textcolor{black}{the deflection at a specific length} along the beam \((m)\), \(L\) is the length of the beam \((m)\) and \(x\) denotes the distance from the fixed end of the beam to a certain point \((m)\). In our design, we focus on the end with the largest deflection. Here \(I\) denotes the moment of inertia in the trace direction \((m^4)\). For the rectangular cross-section cantilever:

\begin{equation}    \label{eq:momentInertia}
    I_T=\frac{1}{24}b^3\left( r+R \right)d\theta
\end{equation}

where \(b\) is the thickness of the sealing lip. According to Equations \ref{eq:areaForce}, \ref{eq:materilMechanics} and \ref{eq:momentInertia}. The deflection (i.e., when \(x = L\)) of the free end \((y)\) can be computed as:

\begin{equation}    \label{eq:diskdeflection}
    \textcolor{black}{y=\frac{3\rho Q^2\left( R-r \right)^3}{b^3E\ (R+r)^2  cos^3\alpha d^2\theta}}
\end{equation}

\subsection{Working Stages} 


The operation of the hybrid gripper involves three stages. \textcolor{black}{The granules first flow upon contact with the object, enabling the gripper to conform to its contours.} As the gripper continues to approach, the granules are compressed against the base. Activating the vacuum then induces a stiffness transition in the granules from a flexible to a rigid state. For small objects, the porous interface may cause insufficient airtight seal \textcolor{black}{that results in consistent airflow and pressure on the sealing lip.} In the last stage, vacuum flow is regulated, enhancing adhesion and airtightness to ensure a secure grasp.

\section{Results}

\subsection{Adaptive Grasping in Various Applications}

\begin{figure}[t]
    \centering
    \includegraphics[width=\columnwidth]{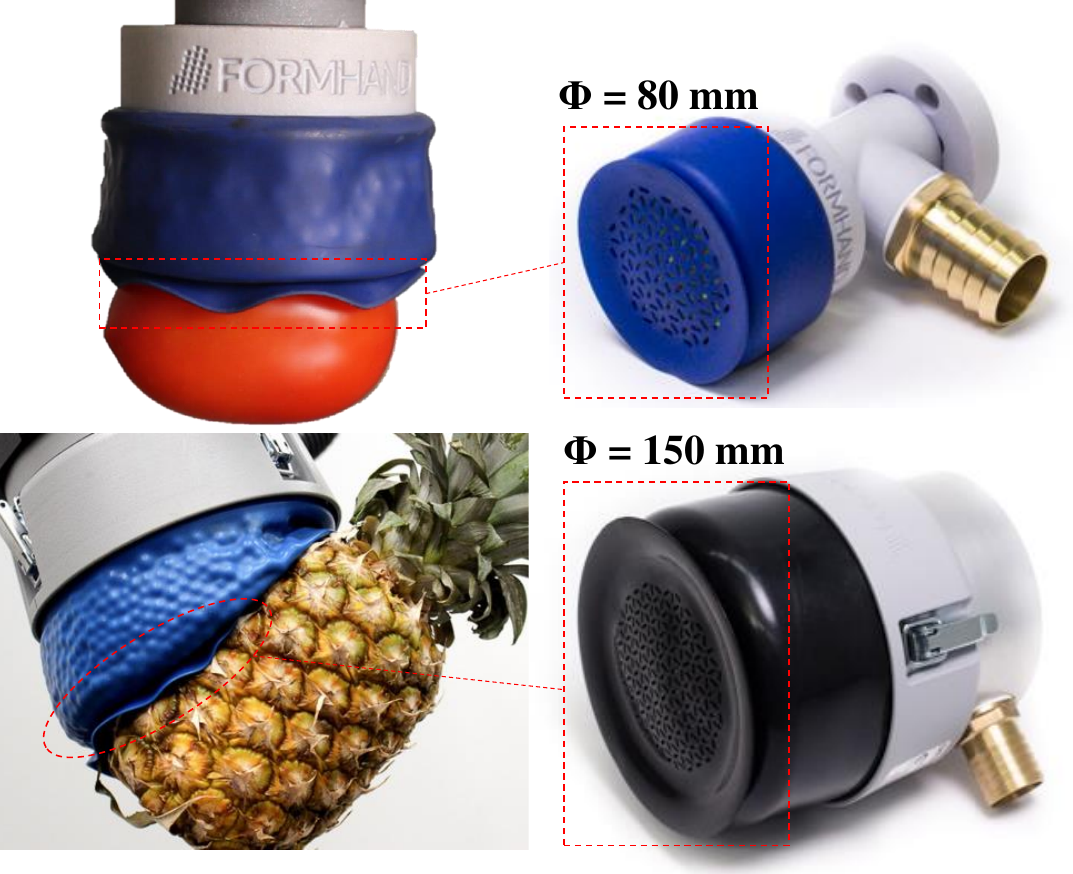}
    \caption{The grasping in real-world applications \textcolor{black}{and the enlarged views of the sealing lips.} (Courtesy of FormHand Automation GmbH.)}
    \label{fig:multi_application}
\end{figure}

Figure \ref{fig:multi_application} shows the gripper's ability to handle a range of objects (see Supporting Video S1). Enhanced performance can be achieved by two or more grippers in an array (Supporting Video S2). This gripper can grasp objects with high load-to-mass ratios \textcolor{black}{such as construction bricks and oxygen tanks.} Figure \ref{fig:grasp6objects}a demonstrates the \textcolor{black}{gripper's versatility in working environments} including the domestic, industrial and specialized tasks such as construction, delicate handling and fruit harvesting. These objects were chosen to \textcolor{black}{represent different grasping conditions} with varying factors like mass, surface rigidity, roughness and shape complexity.

\subsection{Grasping Performance}

\subsubsection{Experimental Setup}

\begin{figure}[b]
    \centering
    \includegraphics[width=\columnwidth]{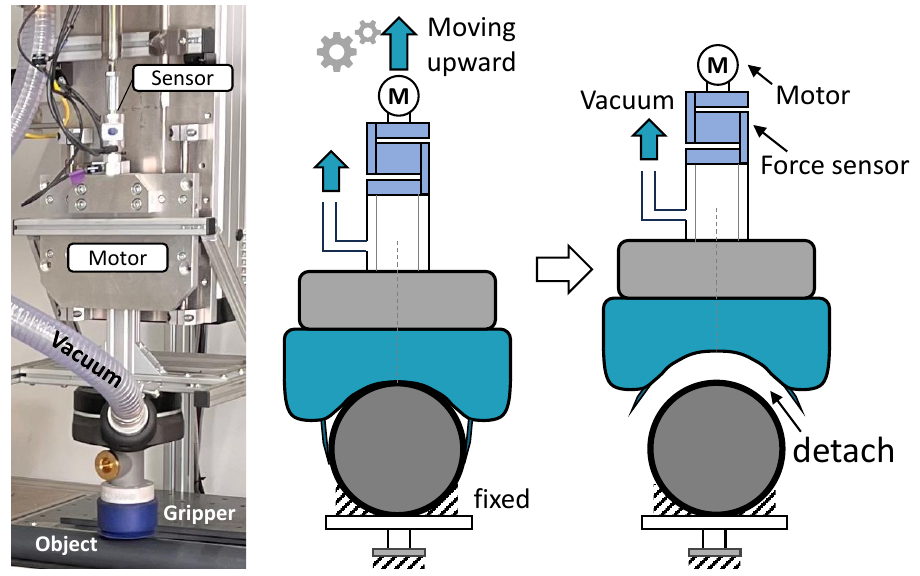}
    \caption{The experimental setup to measure the maximum holding force (MHF).}
    \label{fig:expSetupMHF}
\end{figure}

Figure \ref{fig:expSetupMHF} illustrates the \textcolor{black}{experimental setup of measuring the holding performance. The object is fixed to the ground and the motor is programmed to move downward until the gripper's interface makes contact with the object. Once the vacuum is activated, the motor lifts the entire system while the force sensor measures the handling force.} The maximum holding force is recorded when the object detaches from the gripper. Each object is tested three times for consistency. The FH-R80 gripper is used in experiments to evaluate its holding force and vacuum flow regulation. The gripper weighs 137.5 grams.


\begin{figure*}[!h]
    \centering
    \includegraphics[width=0.93\textwidth]{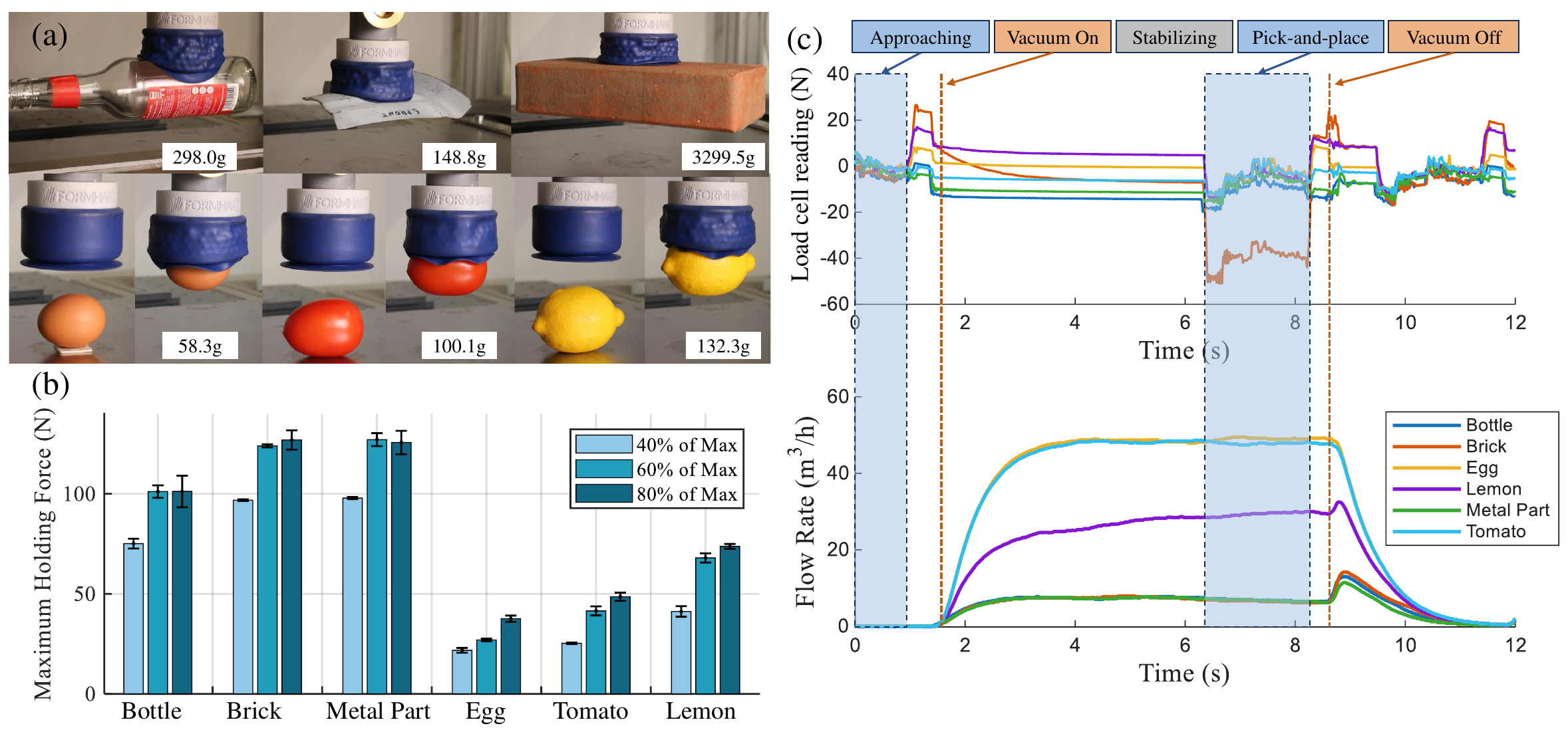}
    \caption{The gripper's holding performance. (a) Grasping various objects. (b) The maximum holding forces (MHFs) versus different objects under different vacuum power conditions. \textcolor{black}{(c) Force sensor readings vs. time (upper) and the flow rate of the gripper (bottom).}}
    \label{fig:grasp6objects}
\end{figure*}

\begin{figure}[h]
    \centering
    \includegraphics[width=0.9\columnwidth]{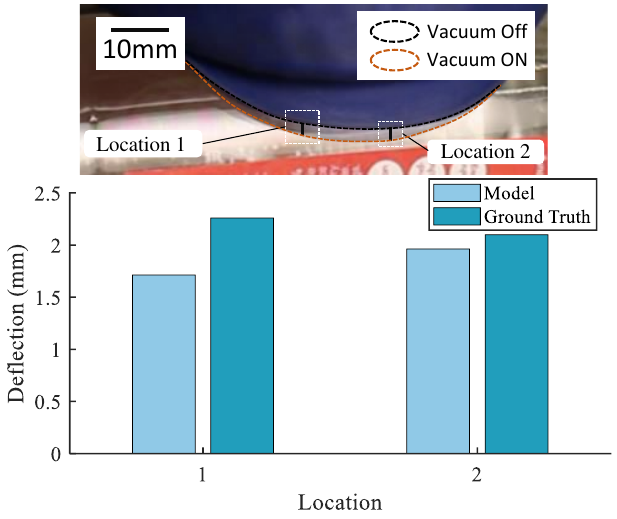}
    \caption{The self-closure of the sealing lip for enhanced adhesion.}
    \label{fig:diskClose}
\end{figure}

\subsubsection{Holding Force}

Figure \ref{fig:grasp6objects}a shows the gripper's capability to safely handle an egg which measures \(54.5\%\) of its aperture, and a 3.3 kg construction brick, achieving a lifting ratio of 24.0 (object mass to gripper mass). The gripper's performance is then quantitatively analyzed under a maximum vacuum blower pressure of 410 mbar, with the power set at \(40\%\). During tests, the gripper moves down by 44 mm, pressing onto and wrapping the object before vacuum activation (Supporting Video S3). This results in either a rigid state (airtight seal) or sealing lip self-closure (incomplete seal). The first row of Figure \ref{fig:grasp6objects}c illustrates the vertical displacement and gripping stages. Under identical vacuum power, the brick and metal part achieve the highest MHF of 127.1 N, while the egg and tomato exhibit similar MHFs, all below 50 N.

\textcolor{black}{This experiment validates the gripper's ability to achieve a load-to-mass ratio of 94.3. While the original particle jamming gripper has shown a maximum ratio of 15.1 \cite{shintake2018soft}, our proposed hybrid design demonstrates an improvement of 5.24 times in handling ability.} The MHF generally increases with vacuum power, with the brick and metal part showing similar MHFs due to comparable adhesion conditions. For the brick, grasping fails at vacuum power below \(40\%\) of the maximum. During stabilization, the egg and tomato exhibit the highest flow rates (\(47.9 \, \text{m}^3/\text{h}\)), followed by the lemon (\(27.5 \, \text{m}^3/\text{h}\)), and the remaining objects (\(7.4 \, \text{m}^3/\text{h}\)). The higher flow rates for the lemon and tomato are due to their smaller diameters relative to the aperture.

\subsubsection{Self-Closure}

The sealing lip's passive adhesion to the object is activated under the vacuum (Supporting Video S4). Using the self-morphing model (Equation \ref{eq:diskdeflection}), we compared the mathematical predictions with the actual grasping behavior. The sealing lip's deflection at two locations closely matches the ground truth, achieving an accuracy of \(84.7\%\). Discrepancies arise mainly from the object's convex shape and the gripper-object orientation.

\section{Conclusion}

This paper proposes a form-flexible design for industrial adaptive grasping. The entire robotic system works with minimized control complexity and information. The design's skirt-shaped sealing lip contributes to the enhanced adhesion under airflow. The reported grippers are suitable for use in warehouses, factories, hospitals and restaurants. Experiments show their ability to grasp objects smaller than their aperture (0.55$\phi$) and handle bulky and heavy items including bricks. The design achieves a maximum load-to-mass ratio of 94.3.




\section*{Supplementary Materials}
The supplementary materials and videos of the proposed design can be viewed at this \href{https://drive.google.com/drive/folders/1442793YUXQFIPUIvfRrW5LjXWSDkA9Js?usp=sharing}{link}.

\section*{Conflict of Interest}
This work is supported by the EU MSCA-ITN project No. 860108. Holger Kunz holds stock ownership in FORMHAND Automation GmbH. The other authors declare no conflict of interest. The authors have applied a Creative Commons Attribution-NonCommercial (CC BY-NC) licence to any Author Accepted Manuscript version arising from this submission.

\bibliographystyle{IEEEtran}
\bibliography{ref}

\begin{thebibliography}{10}
\providecommand{\url}[1]{#1}
\csname url@samestyle\endcsname
\providecommand{\newblock}{\relax}
\providecommand{\bibinfo}[2]{#2}
\providecommand{\BIBentrySTDinterwordspacing}{\spaceskip=0pt\relax}
\providecommand{\BIBentryALTinterwordstretchfactor}{4}
\providecommand{\BIBentryALTinterwordspacing}{\spaceskip=\fontdimen2\font plus
\BIBentryALTinterwordstretchfactor\fontdimen3\font minus \fontdimen4\font\relax}
\providecommand{\BIBforeignlanguage}[2]{{%
\expandafter\ifx\csname l@#1\endcsname\relax
\typeout{** WARNING: IEEEtran.bst: No hyphenation pattern has been}%
\typeout{** loaded for the language `#1'. Using the pattern for}%
\typeout{** the default language instead.}%
\else
\language=\csname l@#1\endcsname
\fi
#2}}
\providecommand{\BIBdecl}{\relax}
\BIBdecl

\bibitem{wang2017prestressed}
Z.~Wang, Y.~Torigoe, and S.~Hirai, ``A prestressed soft gripper: design, modeling, fabrication, and tests for food handling,'' \emph{IEEE Robotics and Automation Letters}, vol.~2, no.~4, pp. 1909--1916, 2017.

\bibitem{shintake2018soft}
J.~Shintake, V.~Cacucciolo, D.~Floreano, and H.~Shea, ``Soft robotic grippers,'' \emph{Advanced materials}, vol.~30, no.~29, p. 1707035, 2018.

\bibitem{wang2023self}
Z.~Wang, S.~Terryn, H.~Wang, J.~Legrand, A.~Safaei, J.~Brancart, G.~V. Assche, and B.~Vanderborght, ``Self-closing and self-healing multi-material suction cups for energy-efficient vacuum grippers,'' \emph{Advanced Intelligent Systems}, vol.~5, no.~10, p. 2300135, 2023.

\bibitem{gilday2020suction}
K.~Gilday, J.~Lilley, and F.~Iida, ``Suction cup based on particle jamming and its performance comparison in various fruit handling tasks,'' in \emph{2020 IEEE/ASME International Conference on Advanced Intelligent Mechatronics (AIM)}.\hskip 1em plus 0.5em minus 0.4em\relax IEEE, 2020, pp. 607--612.

\bibitem{deimel2016novel}
R.~Deimel and O.~Brock, ``A novel type of compliant and underactuated robotic hand for dexterous grasping,'' \emph{The International Journal of Robotics Research}, vol.~35, no. 1-3, pp. 161--185, 2016.

\bibitem{amend2012positive}
J.~R. Amend, E.~Brown, N.~Rodenberg, H.~M. Jaeger, and H.~Lipson, ``A positive pressure universal gripper based on the jamming of granular material,'' \emph{IEEE transactions on robotics}, vol.~28, no.~2, pp. 341--350, 2012.

\bibitem{hwang2023self}
H.~Wang, S.~Terryn, Z.~Wang, G.~Van~Assche, F.~Iida, and B.~Vanderborght, ``Self-regulated self-healing robotic gripper for resilient and adaptive grasping,'' \emph{Advanced Intelligent Systems}, p. 2300223, 2023.

\bibitem{kim2023octopus}
J.~M. Kim, A.~Coutinho, Y.~J. Park, and H.~Rodrigue, ``Octopus-inspired suction cup array for versatile grasping operations,'' \emph{IEEE Robotics and Automation Letters}, vol.~8, no.~5, pp. 2962--2969, 2023.

\bibitem{brown2010universal}
E.~Brown, N.~Rodenberg, J.~Amend, A.~Mozeika, E.~Steltz, M.~R. Zakin, H.~Lipson, and H.~M. Jaeger, ``Universal robotic gripper based on the jamming of granular material,'' \emph{Proceedings of the National Academy of Sciences}, vol. 107, no.~44, pp. 18\,809--18\,814, 2010.

\bibitem{gilday2023xeno}
K.~Gilday, R.~Hashem, A.~Abdulali, and F.~Iida, ``The xeno-tongue gripper: Granular jamming suction cup with bellow-driven self-morphing,'' in \emph{2023 IEEE International Conference on Soft Robotics (RoboSoft)}.\hskip 1em plus 0.5em minus 0.4em\relax IEEE, 2023, pp. 1--6.

\bibitem{fang2022soft}
B.~Fang, Z.~Xia, F.~Sun, Y.~Yang, H.~Liu, and C.~Fang, ``Soft magnetic fingertip with particle-jamming structure for tactile perception and grasping,'' \emph{IEEE Transactions on Industrial Electronics}, vol.~70, no.~6, pp. 6027--6035, 2022.

\bibitem{krahn2017soft}
J.~M. Krahn, F.~Fabbro, and C.~Menon, ``A soft-touch gripper for grasping delicate objects,'' \emph{IEEE/ASME Transactions on Mechatronics}, vol.~22, no.~3, pp. 1276--1286, 2017.

\bibitem{chin2020multiplexed}
L.~Chin, F.~Barscevicius, J.~Lipton, and D.~Rus, ``Multiplexed manipulation: Versatile multimodal grasping via a hybrid soft gripper,'' in \emph{2020 IEEE International Conference on Robotics and Automation (ICRA)}.\hskip 1em plus 0.5em minus 0.4em\relax IEEE, 2020, pp. 8949--8955.

\bibitem{li2023bioinspired}
H.~Li, X.~Li, P.~Zhou, and J.~Yao, ``Bioinspired soft tube-foot array with variable stiffness: Design, characterization, and application,'' \emph{IEEE/ASME Transactions on Mechatronics}, 2023.

\bibitem{lee2021soft}
J.~Lee, J.~Kim, S.~Park, D.~Hwang, and S.~Yang, ``Soft robotic palm with tunable stiffness using dual-layered particle jamming mechanism,'' \emph{IEEE/ASME Transactions on Mechatronics}, vol.~26, no.~4, pp. 1820--1827, 2021.

\bibitem{fang2022multimode}
B.~Fang, F.~Sun, L.~Wu, F.~Liu, X.~Wang, H.~Huang, W.~Huang, H.~Liu, and L.~Wen, ``Multimode grasping soft gripper achieved by layer jamming structure and tendon-driven mechanism,'' \emph{Soft Robotics}, vol.~9, no.~2, pp. 233--249, 2022.

\bibitem{washio2022design}
S.~Washio, K.~Gilday, and F.~Iida, ``Design and control of a multi-modal soft gripper inspired by elephant fingers,'' in \emph{2022 IEEE/RSJ International Conference on Intelligent Robots and Systems (IROS)}.\hskip 1em plus 0.5em minus 0.4em\relax IEEE, 2022, pp. 4228--4235.

\bibitem{hao2020multimodal}
Y.~Hao, S.~Biswas, E.~W. Hawkes, T.~Wang, M.~Zhu, L.~Wen, and Y.~Visell, ``A multimodal, enveloping soft gripper: Shape conformation, bioinspired adhesion, and expansion-driven suction,'' \emph{IEEE Transactions on Robotics}, vol.~37, no.~2, pp. 350--362, 2020.

\bibitem{lochte2014form}
C.~L{\"o}chte, H.~Kunz, R.~Schnurr, S.~Langhorst, F.~Dietrich, A.~Raatz, K.~Dilger, and K.~Dr{\"o}der, ``Form-flexible handling and joining technology (formhand) for the forming and assembly of limp materials,'' \emph{Procedia CIRP}, vol.~23, pp. 206--211, 2014.

\end{thebibliography}

\balance

\balance

\end{document}